\documentclass[11pt]{article}

\usepackage[]{EMNLP2023}

\usepackage{times}
\usepackage{latexsym}
\usepackage{epsfig}
\usepackage{graphicx}
\usepackage{amsmath}
\usepackage{amssymb}
\usepackage{multirow}
\usepackage{spverbatim}
\usepackage{xcolor}
\usepackage{xspace}
\usepackage{cleveref}
\usepackage{makecell}

\crefformat{section}{\S#2#1#3} 
\crefformat{subsection}{\S#2#1#3}
\crefformat{subsubsection}{\S#2#1#3}

\usepackage[T1]{fontenc}

\usepackage[utf8]{inputenc}

\usepackage{microtype}

\usepackage{inconsolata}

\newcommand\vlistitlefont[1]{\textbf{\smash{\fontfamily{cmtt}\selectfont#1}}}

\definecolor{sgreen}{HTML}{39C277}
\definecolor{sblue}{HTML}{0070C0}
\definecolor{sred}{HTML}{FF0000}

\makeatletter
\DeclareRobustCommand\onedot{\futurelet\@let@token\@onedot}
\def\@onedot{\ifx\@let@token.\else.\null\fi\xspace}

\def\eg{\emph{e.g}\onedot}

\newcommand*{\shifttext}[2]{%
  \settowidth{\@tempdima}{#2}%
  \makebox[\@tempdima]{\hspace*{#1}#2}%
}
\makeatother

\newcommand{\logo}[0]{\text{\smash{\raisebox{-1pt}{\includegraphics[height=8pt]{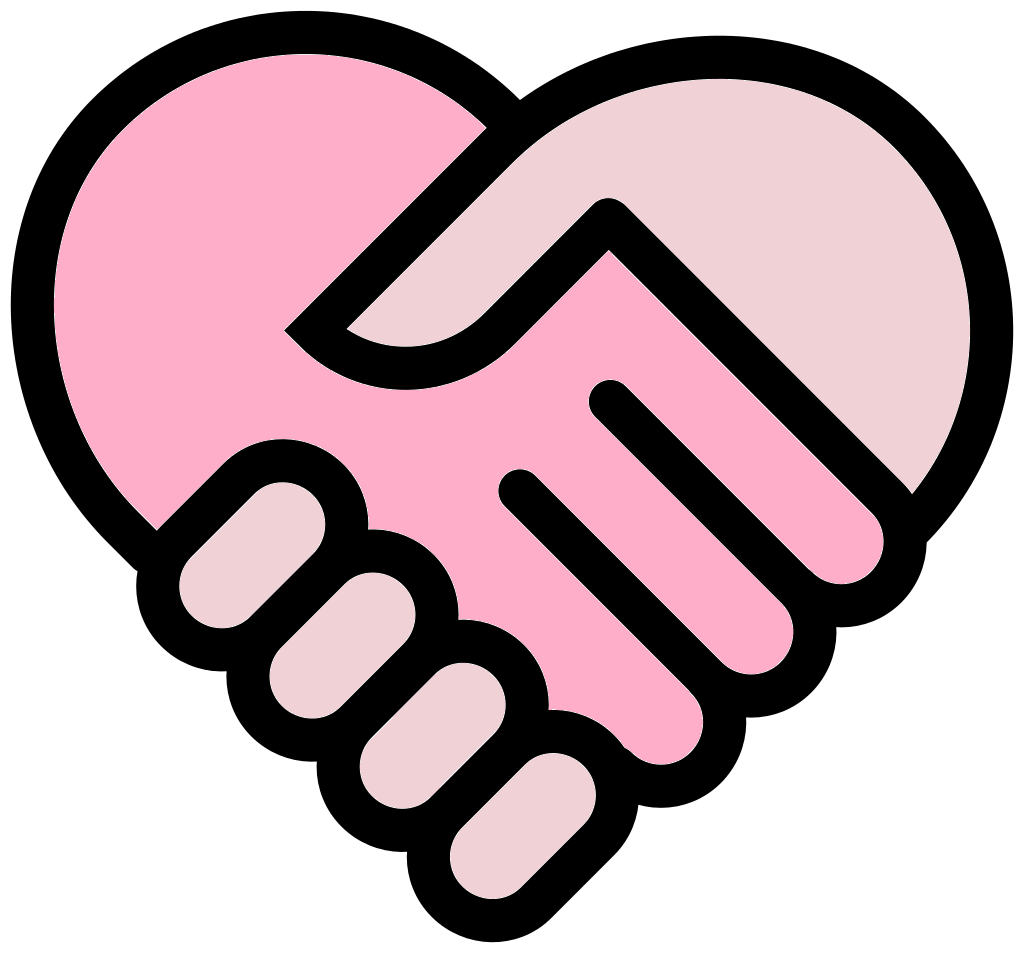}}}}\xspace}
\newcommand{\logolarge}[0]{\text{\smash{\raisebox{-2pt}{\includegraphics[height=13pt]{graphics/logo.png}}}}\xspace}

\newcommand{\modelnamelong}{Visual-Language models as Importance Sampling weights\xspace}
\newcommand{\modelnametitle}{\shifttext{2pt}{\logolarge} \vlistitlefont{VLIS}\xspace}
\newcommand{\modelnamefancy}{\shifttext{2pt}{\logo} \vlistitlefont{VLIS}\xspace}
\newcommand{\modelname}{\vlistitlefont{VLIS}\xspace}
\newcommand{\naive}{Na\"ive\xspace}

\title{\modelnametitle: Unimodal Language Models Guide\\Multimodal Language Generation}

\author{
    Jiwan Chung \and  Youngjae Yu \\
    Yonsei University \\
    \\
    \url{https://github.com/JiwanChung/vlis}
}

\begin{document}
\maketitle
\begin{abstract}
Multimodal language generation, which leverages the synergy of language and vision, is a rapidly expanding field.
However, existing vision-language models face challenges in tasks that require complex linguistic understanding.
To address this issue, we introduce \modelnamelong (\modelnamefancy), a novel framework that combines the visual conditioning capability of vision-language models with the language understanding of unimodal text-only language models without further training.
It extracts pointwise mutual information of each image and text from a visual-language model and uses the value as an importance sampling weight to adjust the token likelihood from a text-only model.
\modelname improves vision-language models on diverse tasks, including commonsense understanding (WHOOPS, OK-VQA, and ScienceQA) and complex text generation (Concadia, Image Paragraph Captioning, and ROCStories). Our results suggest that \modelname represents a promising new direction for multimodal language generation.
\end{abstract}

\section{Introduction}
\label{sec:introduction}

Visual Language Models (VLMs) extend unimodal text-only language models by conditioning their outputs on image context.
Recent VLMs ~\cite{li2022blip,li2023blip,wang2022git} can perform diverse multimodal tasks from commonsense VQAs ~\cite{okvqa2019cvpr,schwenk2022okvqa} to in-context learning~\cite{alayrac2022flamingo,awadalla2023openflamingo,huang2023kosmos}.
Moreover, instruction tuning with visual inputs~\cite{liu2023llava,li2023otter,instructblip} has improved the VLMs' responsiveness to an even more extensive variety of tasks~\cite{lu2022scienceqa,yang2021just}.

\begin{figure}[ht]
    \centering
    \includegraphics[width=0.48\textwidth]{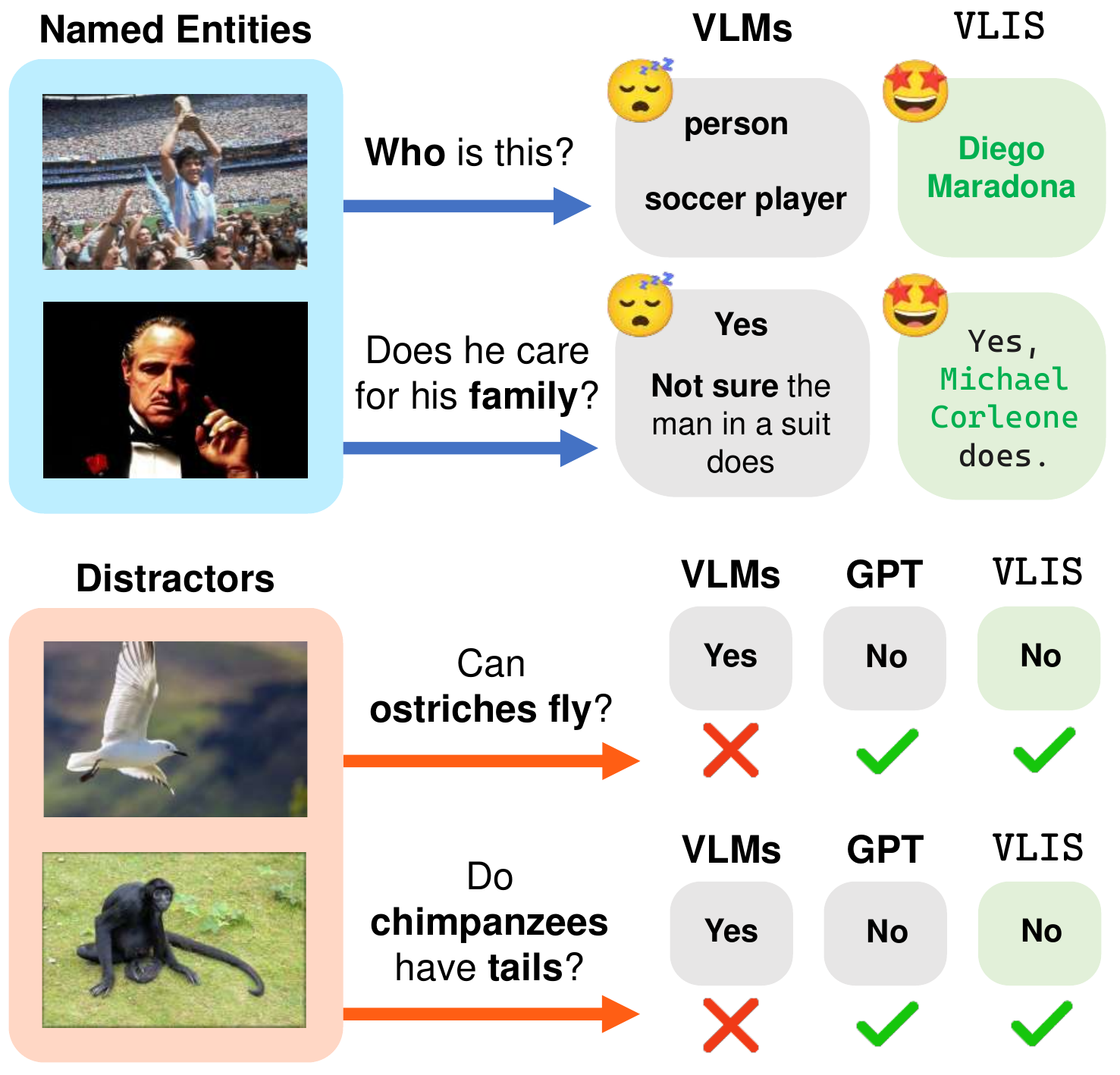}
    \caption{\textbf{TOP}: \modelname correctly recognizes named entities,
    unlike the VLMs. \textbf{Bottom}: \modelname is not deceived by the distractor images. Note that the images show a seagull and a monkey, not an ostrich and a chimpanzee.
    \modelname inherits this favorable linguistic capability from a text-only language model~\cite{Touvron2023LLaMAOA,zhang2022opt},
    and use VLMs as a guide for visual alignment.
    The examples are truncated for visualization purposes:
    we provide the full-text in~\cref{subsec:ax_failure_samples}.
    }
    \label{fig:main_intuition}
\end{figure}

However, most VLMs 
only partially inherit 
the linguistic understanding capability
of the unimodal models~\cite{iki2021effect}.
We here illustrate two intriguing failure cases of the recent VLMs,
using both a strong image captioning model (BLIP-2~\cite{li2023blip})
and an instruction-tuned model (LLAVA~\cite{liu2023llava}).
Firstly, VLMs avoid specifying named entities. 
The upper examples of Figure~\ref{fig:main_intuition} show the VLM failing to describe a public figure (\textit{Diego Maradona}) or movie character (\textit{Don Corleone}).
The problem is not the lack of knowledge: after applying our zero-shot method (\modelname), the VLM tells the names.
We further investigate this phenomenon in the landmark recognition experiment in \cref{subsec:ax_landmark}.

\begin{figure*}[ht]
    \centering
    \includegraphics[width=0.98\textwidth]{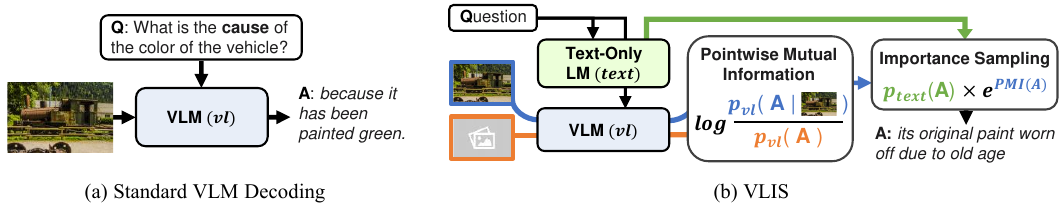}
    \caption{
    Comparison of \modelname and standard VLM decoding process.
    Using the VLM, we first obtain the image-conditional $p_{vl}(Answer|image)$ and text-only likelihood $p_{vl}(Answer)$ given an image and a prompt or question.
    Then, we compute the exponentiated pointwise mutual information (PMI) with the likelihoods.
    Finally, the exponentiated PMI score is used as the importance weights for the text-only model likelihood $p_{text}(Answer)$.
    }
    \label{fig:main_architecture}
\end{figure*}

Secondly, VLMs rely on the image context, even when they should not.
The lower examples of the same figure show the VLM being misled by
image context to deny commonsense knowledge.
The questions are not unanswerable: the text-only language model without the image context answers both correctly.
We provide more samples on visual distraction in \cref{subsec:ax_failure_samples}.

Hence, the linguistic capabilities of the VLMs are not optimal yet.
On the other hand, the unimodal text-only language models themselves~\cite{brown2020gpt3,Touvron2023LLaMAOA}
show reliable linguistic understanding
and known for
their knowledge understanding~\cite{petroni2019language,meng2022locating}
and complex reasoning capabilities~\cite{Kojima2022LargeLM,qin2023chatgpt}.
Hence, it becomes reasonable to
delegate the burden of language modeling to the text-only models.

To this end, we propose \modelnamelong (\modelnamefancy)
as a plug-and-play method to
enhance the unreliable linguistic understanding of the VLMs.
When generating each text token, \modelname follows the token likelihoods of the unimodal \textbf{text-only} language model.
Furthermore, \modelname multiplies importance sampling~\cite{tokdar2010importance} weights derived from a VLM to provide the \textbf{visual alignment} signals.
To isolate the visual conditioning capability of the VLMs
from their language modeling preference, we incorporate the exponentiated pointwise mutual information (PMI)~\cite{church1990pmi}
of the image context and the current text token
as the weights.
As a result, \modelname can maintain the favorable language modeling capability of the text-only model and control the visual conditioning strength simultaneously.

We evaluate \modelname on two VLM backbones
to test whether \modelname is effective
both when the language modeling capability of the VLM is weaker
than that of the text-only model
(BLIP-2~\cite{li2023blip})
and when the VLM is expected to model language well
owing to the visual instruction tuning process
(LLAVA~\cite{liu2023llava}).
Our experiments consist of various tasks that require
both reliable language modeling and strong visual conditioning,
including weirdness identification (WHOOPS~\cite{bitton2023whoops}) and commonsense VQA (OK-VQA~\cite{okvqa2019cvpr}, ScienceQA~\cite{lu2022scienceqa}), 
extended image captioning
(Concadia~\cite{kreiss2022concadia} and Image Paragraph Captioning~\cite{krause2017imageparagraph}), and open-ended generation (ROCStories~\cite{mostafazadeh2016rocstories}).
Compared to the dataset-specific state-of-the-art baselines and the base VLMs,
\modelname improves linguistic capabilities such as responsiveness to prompts 
 while maintaining visual conditioning according to a comprehensive set of evaluation metrics.

\section{VLMs as Importance Sampling Weights}
\label{sec:method}

We propose \modelnamelong (\modelname) to harmonize the visual conditioning capability of the VLMs with the linguistic fluency of the text-only language models.
We provide the intuition behind our approach in \cref{subsec:method_intuition},
describe our token-level visual alignment scores in \cref{subsec:method_pmi},
and combine the said scores with the text-only model via importance sampling in \cref{subsec:method_importance_sampling}.

\subsection{Intuition}
\label{subsec:method_intuition}

Many recent Visual Language Models (VLMs)~\cite{li2023blip,alayrac2022flamingo,liu2023llava} are often
built on top of text-only language models~\cite{iyer2022opt,hoffmann2022chinchilla,Touvron2023LLaMAOA}.
At each timestep $t$,
the per-token likelihood of the autoregressive text-only language models is modeled as $p_{text} (x_t | x_{<t})$, where $x$ denotes a text token.
To build a VLM $p_{vl}$, one can finetune the text-only model on data $S$ consisting of paired image $c$ and text $x$ with maximum likelihood estimation as the objective.
\begin{align}
    \theta_{vl} &\sim argmin_{\theta} E_{(x, c) \in S} [-\log p_{\theta} (x | c) ]  \label{eq:1} 
\end{align}

However, while this objective only maximizes the image-conditional likelihood $p_{vl} (x_t|c)$,
it may lead to unpredicted artifacts in the marginal likelihood $p_{vl} (x_t)$ that does not depend on any particular image.
For example, image captioning models are known to not only reflect but also amplify the social bias present in the training data~\cite{hendricks2018women},
or distort the original language model's commonsense knowledge as described in \cref{sec:introduction}.

We henceforth seek to extract the visual conditioning capability of the VLMs isolated from their dubious language modeling skills.

\subsection{Extracting Visual Weights}
\label{subsec:method_pmi}

Here, we aim to find a quantity that extracts the visual conditioning strength of a VLM stripped of its language modeling preference.
We employ Pointwise Mutual Information (PMI)~\cite{church1990pmi}, which measures the association between two events (text and image in our setting).
On each step, we want to compute the PMI between the image context $c$ and the next token $x_t$ given the previous text context $x_{<t}$:
\begin{align}
    PMI(x_t| c, x_{<t}) &= \
    \log \frac{p_{vl}(x_t, c | x_{<t})}{p_{vl}(x_t | x_{<t}) p_{vl}(c)} 
    \label{eq:3} \\
    &= \
    \log \frac{p_{vl}(x_t | c, x_{<t})}{p_{vl}(x_t | x_{<t})}
    \label{eq:4} 
\end{align}

\cref{eq:4} reformulates the definition in \cref{eq:3} for better tractability.
The numerator is the image-conditional likelihood of the VLM
and is easily obtainable.
However, the denominator requires marginalization over the image context $c$.
We enumerate three proxies below that bypass the excessive computation required to obtain the expectation over all possible images.

\textbf{Approximating the marginal}.
The first approximation is training a separate text-only model with the VLMs' training data $S$.
Considering the massive scale of dataset $S$, this option requires a considerable burden of additional training.
Also, there is no guarantee that the newly trained model will accurately estimate the marginal likelihood due to the additional complexity of training another model.
The second option is using a sample mean of the pre-selected image set as a proxy to the real mean.
Lastly, the score for only one or two images might suffice as the sample image set. 

We use the last method with the least computational overhead.
Here, the sample set is a tiny set of images with close to no visual information. In practice, we use two images: a black-filled image $c_b$ and a white-filled image $c_w$.
\begin{align}
    p_{vl} (x_t | x_{<t})
    \sim \frac{1}{2} \sum_{c \in [c_b, c_w]}
    &p_{vl} (x_t | x_{<t}, c) 
\end{align}

This efficient alternative works reasonably well in practice and is used in all our experiments.
As a result, \modelname runs three forward passes of VLM (one for the conditional likelihood and two for the marginal likelihood) and a single pass of the text-only model on each step of the generation process.
We explore more choices of selecting the marginal image set later in~\cref{sec:ax_marginal}, which shows that our specific set of images provides a reasonable trade-off between generation quality and inference time.

\subsection{Computing \modelname Scores}
\label{subsec:method_importance_sampling}

We start from the token likelihood of text-only language models $p_{text} (x_t|c, x_{<t})$.
To control the degree of confidence in the text-only models' decisions,
we introduce a language temperature $\tau$ to smooth or de-smooth the text-only distributions:
\begin{align}
    \bar{p}_{text}(x_t|c, x_{<t})
    &\propto 
    p_{text}(x_t|c, x_{<t})^{\frac{1}{\tau}}
\end{align}

Then, we multiply the exponentiated PMI introduced in \cref{subsec:method_pmi} with the likelihood for better visual alignment.
\modelname decides the next token $x_t$ with the following score $f(x_t)$:
\begin{align}
    f(x_t)
    &= \bar{p}_{text}(x_t|c, x_{<t}) e^{PMI(x_t, c|x_{<t}))} \\
    &= \bar{p}_{text}(x_t|c, x_{<t})
    \frac{p_{vl}(x_t | c, x_{<t})}{p_{vl}(x_t | x_{<t})} \label{eq:8}
\end{align}

Written as \cref{eq:8}, \modelname performs importance sampling of the smoothed text-only model likelihood $p_{text}$.
Importance sampling~\cite{tokdar2010importance} is a Monte-Carlo method of estimating a quantity $v(x)$
from the \textit{nominal distribution} $p(x)$ with samples from another distribution called \textit{importance distribution} $q(x)$.
The estimated quantity here is the text-only model likelihood $\bar{p}_{text}(x_t)$,
the nominal distribution is the VLMs' image-conditioned likelihood $p_{vl}(x_t|c)$,
and the importance distribution is the marginal $p_{vl}(x_t)$.
\begin{align}
    E[f(x_t):P] &\sim E_{x_t \sim q(x_t)} [v(x_t) \frac{p(x_t)}{q(x_t)}] \\
    v(x_t) &:= \bar{p}_{text} (x_t | x_{<t}) \nonumber \\
    p(x_t) &:= p_{vl} (x_t | c, x_{<t}) \nonumber \\
    q(x_t) &:= p_{vl} (x_t | x_{<t}) \nonumber
\end{align}

Implementation-wise, we replace the expectation with a single sample (current generated text).
Thus, \modelname effectively treats the current token candidate as sampled from
the VLMs' marginal $p_{vl}(x_t)$ and reweigh its importance with the VLMs' conditional $p_{vl}(x_t|c)$.


\textbf{Fluency masking}.
The log visual weights $PMI(x_t,c|x_{<t})$ of \modelname is a log-likelihood ratio and is unbounded.
Hence, some extreme cases, such as tiny values of the marginal likelihood $p_{vl} (x_t|x_{<t})$ may overrule the language generation process of the text-only model, yielding degenerate text.
To prevent such text degeneracy, we apply a fluency mask to our importance sampling score $f(x_t | x_{<t}, c)$: only the tokens with text-only likelihood larger than the threshold $\alpha$ are allowed to be selected. We omit the dependency on the contexts $x_{<t},c$ in the equation below for simplicity.
\begin{align}
    \tilde{f} (x_t) = 
    \begin{cases}
        f (x_t), & if\  x_t \in \mathcal{V}_{top}\\
        -inf, & otherwise
    \end{cases} \\
    \mathcal{V}_{top} = \{ x_t | p_{text} (x_t) \ge \alpha \}
\end{align}

Intuitively, this mask filters out any token candidates the text-only model sees as the next token with a probability lower than $\alpha$.
We fix the fluency threshold to $\alpha=0.001$ in all experiments except for an alternative architecture (\cref{sec:ax_backbones}). 
Still, \modelname is not overly sensitive to the specific value of the fluency threshold. We conduct a hyperparameter search experiment to verify this in \cref{sec:ax_fluency_threshold}.

The token that maximizes this final score $\tilde{f} (x_t|c,x_{<t})$ is greedily selected as the next token. When \modelname is combined with other decoding methods, such as beam search, the score substitutes the original token likelihood as per-token scores.

\section{Experiments: Describing Facts}
\label{sec:exp_factual}

We verify that \modelname can alleviate the factual inaccuracy concern raised
in Figure~\ref{fig:main_intuition} with various multimodal benchmarks:
weirdness identification~\cref{subsec:exp_whoops}, commonsense understanding~\cref{subsec:exp_vqa},
and scientific reasoning~\cref{subsec:exp_sqa}.
\modelname consistently outperforms the backbone VLM and
shows comparable factual correctness to the strong baselines.

\textbf{Experimental setups}.
We explore two experimental setups.
Our experiments on the WHOOPS dataset incorporate LLAVA~\cite{liu2023llava} and
Lynx~\cite{zeng2023lynx}
as the VLMs and Vicuna 7B~\cite{vicuna2023} as the text-only model.
In the VQA experiments, we use BLIP-2 OPT 2.7B~\cite{li2023blip} and OPT IML Max 1.3B~\cite{iyer2022opt} as our backbones.\footnote{
We assign different tasks for different backbones for fair comparisons:
BLIP-2 fails to generate long explanation-of-violation since it is only trained on short captions, while it is not trivial to evaluate LLAVA on short-answer VQAs in a zero-shot manner due to its tendency to generate long explanations.
}
Note that the choices of model pairs are intentional: we impose similar computational requirements on both the VLM and the text-only model to limit the additional computational burden of \modelname.
In both cases, we use the base VLM as a general baseline to evaluate the gain from \modelname.
Also, to verify the contribution of the PMI weights, we implement \naive Ensemble which simply multiplies the token likelihood of the VLM and the text-only model.

\textbf{Evaluation metrics}.
We evaluate closed-ended questions with binary (WHOOPS) and multi-choice (ScienceQA) accuracy.
The open-ended VQAs (OK-VQA and VQAv2) use the task-specific VQA metric~\cite{antol2015vqa_iccv}.

\subsection{Identification of Weird Images}
\label{subsec:exp_whoops}

\begin{table}[t]
    \centering
    \begin{tabular}{l|cc|c}
        Models & Pipe & 0-shot  & Acc (\%) \\
        \hline
        Chance & &  & 50 \\
        \hline
        BLIP-2 & & $\checkmark$ & 50 \\
        BLIP-2 & & & \underline{73} \\
        Model Caption & $\checkmark$ & $\checkmark$ & 59 \\
        GT Caption & $\checkmark$ & $\checkmark$ & \underline{74} \\
        VLM (LLAVA) & $\checkmark$ & $\checkmark$ & 59 \\
        VLM (Lynx) & $\checkmark$ & $\checkmark$ & 71 \\
        \hline
        Ours (LLAVA) & $\checkmark$ & $\checkmark$ & \underline{73} \\
        Ours (Lynx) & $\checkmark$ & $\checkmark$ & \textbf{80}
    \end{tabular}
    \caption{Results in the \textit{identification of weird images} task of WHOOPS dataset~\cite{bitton2023whoops}.
    \textit{Pipe} represents further pipelining with GPT3 and \textit{0-shot} denotes a zero-shot method.
    The best numbers are \textbf{bolded} and the second best ones are \underline{underlined}.
    }
    \label{tab:whoops_table}
\end{table}

\begin{figure}
    \centering
    \includegraphics[width=0.48\textwidth]{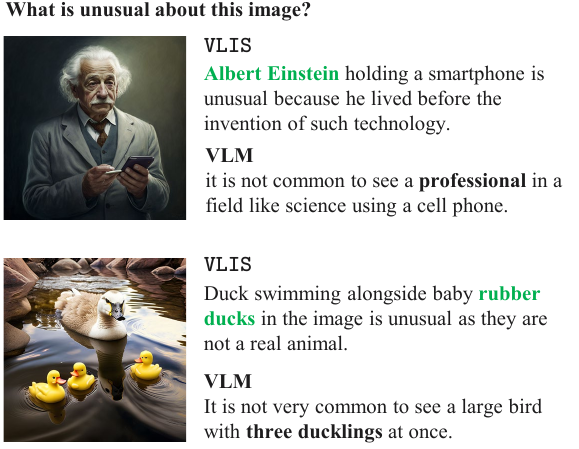}
    \caption{
    Qualitative samples from WHOOPS~\cite{bitton2023whoops} experiments.
    As marked in
     \textbf{\textcolor{sgreen}{green}},
     specific descriptions are required to explain weirdness.
    }
    \label{fig:main_whoops}
\end{figure}

WHOOPS~\cite{bitton2023whoops} is a visual commonsense benchmark to check a VLM's capability to understand images that defy commonsense.
We adopt \textit{identification of weird images},
a subtask of the WHOOPS benchmark,
which tasks a model to discriminate potentially weird images.

\textbf{Approach and Baselines}.
Following the original paper~\cite{bitton2023whoops}, we employ \textit{pipelining} to turn
the original binary classification problem into a description generation problem.
Specifically, \textit{pipelining} means that a model first generates
explanation-of-violation (EoV) description of the given two images,
which is then processed to the off-the-shelf text-only classifier GPT3~\cite{brown2020gpt3} to yield a binary decision on which image is weird.
We use \modelname to generate such EoV descriptions.
The pipelined baselines include EoV from the backbone VLM (LLAVA), 
conventional machine-generated captions, and
ground-truth captions from the WHOOPS dataset.
We also include pipeline-less BLIP-2 (both supervised and zero-shot)
as a baseline.
The same prompt we used for both \modelname and the backbone VLM is illustrated in~\cref{sec:ax_templates}.

\textbf{Results}.
Table~\ref{tab:whoops_table} and Figure~\ref{fig:main_whoops} presents results with LLAVA~\cite{liu2023llava},
an instruction-tuned VLM.
\modelname-generated weirdness explanations perform on par with
the ground-truth captions, which are manually annotated to contain details
necessary to identify the strangeness.
Also, our method as a zero-shot method shows comparable performance to the supervised baseline BLIP-2.
Interestingly, LLAVA alone cannot outperform conventional captions, even with instruction tuning and prompting.

\subsection{Commonsense Understanding}
\label{subsec:exp_vqa}

\begin{table}[t]
    \centering
    \begin{tabular}{p{1.8cm}|cc|cc}
        Models & V & L & OKVQA & VQAv2 \\
        \hline
        FewVLM & \checkmark & & 16.5 & 47.7 \\
        Frozen & \checkmark & & 5.9 & 29.6 \\
        VLKD & \checkmark & & 13.3 & 42.6 \\
        \hline
        BLIP-2 & \checkmark & & 31.7 & 53.5  \\
        OPT-IML & & \checkmark & 19.1 & 36.0 \\
        \multirowcell{2}[0pt][l]{
        \naive \\ Ensemble}
        & \multirow{2}{*}{\checkmark}
        & \multirow{2}{*}{\checkmark}
        & \multirow{2}{*}{26.6}
        & \multirow{2}{*}{34.6} \\
        &&&& \\
        \hline
        Ours & \checkmark & \checkmark  & \textbf{34.2} & \textbf{53.6} 
    \end{tabular}
    \caption{Results in the validation set of OK-VQA~\cite{okvqa2019cvpr} and VQAv2~\cite{goyal2017vqav2}. \textit{V} denotes using a VLM and \textit{L} denotes using a unimodal language model.}
    \label{tab:okvqa_table}
\end{table}
\begin{table}
    \centering
    \begin{tabular}{l|ccc|c}
        Models & IMG & TXT & NO & ALL \\
        \hline
        UnifiedQA\textsubscript{Small} & 44.1 & 50.2 & 44.5 & 45.8 \\
        UnifiedQA\textsubscript{Base} & 48.1 & 53.1 & 46.7 & 48.5 \\
        GPT-3  & 65.7 & 74.2 & 79.6 & 74.0 \\
        \hline
        BLIP-2 & 35.5 & 34.6 & 24.2 & 28.2 \\
        OPT-IML & 45.4 & 52.2 & 49.8 & 49.0 \\
        \naive Ensemble & 45.9 & 53.6 & 49.7 & 49.7\\
        \hline
        Ours & \textbf{49.3} & 53.1 & 49.1 & \textbf{50.2}
    \end{tabular}
    \caption{Zero-shot results on ScienceQA test set~\cite{lu2022scienceqa}.
    \textit{IMG} denotes subset with image context, \textit{TXT} the text context subset,
    and \textit{NO} the subset without any context.
    }
    \label{tab:science_qa_table}
\end{table}
Unimodal language models embody commonsense knowledge~\cite{petroni2019language,davison2019commonsense,tamborrino2020pre}.
If \modelname can inherit this commonsense understanding capability,
it would outperform the base VLM in tasks requiring both commonsense and visual understanding.
Here, we examine this possibility with a commonsense VQA benchmark of OK-VQA~\cite{okvqa2019cvpr}. Further, \modelname is also shown to maintain visual specificity
in VQAv2~\cite{goyal2017vqav2}.

\textbf{Approach and baselines}.
We use OK-VQA~\cite{okvqa2019cvpr} as an example of commonsense-augmented VQA and VQAv2~\cite{goyal2017vqav2} as a visually intensive VQA problem.
We compare \modelname with strong VLM models, including
FewVLM~\cite{jin2022fewvlm}, Frozen~\cite{tsimpoukelli2021frozen},
and VLKD~\cite{dai2022vlkd}.

\textbf{Results: commonsense knowledge}.
In the OK-VQA~\cite{okvqa2019cvpr} experiment in Table~\ref{tab:okvqa_table}, we show that
\modelname achieves meaningful development over the backbone VLM (BLIP-2).
Also, the text-only backbone (OPT-IML) and \naive Ensemble perform substantially worse,
proving that 
\modelname is not just imitating the text-only model outputs.
Instead, \modelname adaptively fuses the commonsense understanding capability of the text-only model with the visual conditioning of the VLM.

\textbf{Results: maintaining visual specificity}.
When VQAs do not require text-based reasoning, 
\modelname should focus on visual conditioning only.
The rightmost column of Table~\ref{tab:okvqa_table}
summarizes results on VQAv2~\cite{goyal2017vqav2} dataset,
a VQA dataset that has its textual bias intentionally removed.
As shown in the VQA score, \modelname (Ours) preserves the VQA capability of the backbone VLM (BLIP-2).
Note that \naive Ensemble falls behind the text-only backbone (OPT-IML), offering a poor trade-off between visual and linguistic understanding.

\subsection{Scientific Reasoning}
\label{subsec:exp_sqa}
ScienceQA~\cite{lu2022scienceqa} evaluates multimodal science reasoning capability. 
Here, the goal of \modelname would be to improve the answers in the presence of image contexts (IMG)
and preserve the answers from the text-only model in the absence of such visual context (TXT and NO).

\textbf{Baselines}.
We compare our zero-shot \modelname against zero-shot baselines 
including a VLM (UnifiedQA~\cite{khashabi2020unifiedqa})
and a text-only language model (GPT-3~\cite{brown2020gpt3}).

\textbf{Results}.
Table~\ref{tab:science_qa_table} demonstrates the findings in ScienceQA.
On IMG split, \modelname significantly improves the text-only OPT-IML and \naive Ensemble baselines. 
Also, \modelname maintains the performance of the text-only backbone
in TXT and NO split.
Finally, the base VLM (BLIP-2) falls behind by a wide margin,
indicating that solid language understanding is necessary for scientific reasoning.

\begin{table}
    \centering
    \begin{tabular}{l|c|c|c}
    Model & Zeroshot & Cap & Desc \\
    \hline
    Kreiss et al. & & 11.3 & 17.4 \\
    Socratic Model & \checkmark & 38.9 & 22.6 \\
    \hline
    BLIP-2 & \checkmark & 20.0 & \textbf{30.6} \\
    \naive Ensemble & \checkmark & 24.7 & 18.4 \\
    \hline
    Ours & \checkmark & \textbf{44.1} & 28.3
    \end{tabular}
    \caption{Results on Concadia~\cite{kreiss2022concadia} test set. \textit{Cap} denotes caption and \textit{Desc} description annotations. We report CIDEr following the literature.}
    \label{tab:concadia_table}
\end{table}
\begin{table}
    \centering
    \begin{tabular}{l|c|c|c|c}
    Model & Shots & M & C & B4 \\
    \hline
    Krause et al. & Full & 16.0 & 13.5 & 8.7 \\
    Liang et al. & Full & 17.1 & 16.8 & 9.0 \\
    SCST & Full & 13.6 & 13.8 & 5.9 \\
    SCST\textsubscript{Rep. Penalty} & Full & 17.9 & 30.6 & 10.6 \\
    HSGED & Full & 18.3 & 36.0 & 11.3 \\
    PaG-MEG-SCST  & Full & 18.2 & 29.4 & 11.5 \\
    \hline
    BLIP-2 & 3 & 10.8 & 6.5 & 4.9 \\
    OPT-IML & 3 & 9.5 & 2.5 & 2.2 \\
    \naive Ensemble & 3 & 9.8 & 6.0 & 3.6 \\
    \hline
    Ours & 3 & \textbf{14.6} & \textbf{14.8} & \textbf{6.4}
    \end{tabular}
    \caption{Results on the Paragraph Captioning~\cite{krause2017imageparagraph} test set.
    \textit{M} denotes METEOR, \textit{C} CIDEr, and \textit{B4} Bleu-4 scores.}
    \label{tab:paracap_table}
\end{table}

\section{Experiments: Text Generation}
\label{sec:exp_cap}

In addition to factual knowledge,
text-only language models manifest two critical capabilities:
 following prompt instructions and 
 generating fluent and diverse text.
 We demonstrate that \modelname extends these qualities to the visual domain
 with contextualized captioning~(\cref{subsec:exp_concadia}),
 paragraph captioning~(\cref{subsec:exp_para}),
 and visual story generation~(\cref{subsec:exp_story}).

\textbf{Metrics}.
Both captioning benchmarks use automatic text metrics, including CIDEr~\cite{vedantam2015cider}, METEOR~\cite{banerjee-lavie-2005-meteor},
and Bleu-4~\cite{papineni-etal-2002-bleu}.
In the open-ended generation problem of visual storytelling,
we use various fluency metrics (2-gram repetition, diversity,
coherence, MAUVE~\cite{pillutla2021mauve}) and
a visual strength metric (CLIPScore~\cite{hessel2021clipscore}).
Refer to \citep{su2022magic} for details on the fluency metrics.

\subsection{Contextualized Captioning}
\label{subsec:exp_concadia}
Concadia~\cite{kreiss2022concadia} is an image captioning dataset with the additional context of a paragraph from the Wikipedia article.
The dataset provides two types of annotations:
\textit{caption}, which takes the article into account
and \textit{description}, which ignores the article context. 

\textbf{Approach and Baselines}.
Following the original evaluation scheme~\cite{kreiss2022concadia},
we generate a single text to compare against both the ground-truth \textit{caption} and \textit{description}.
We include both supervised~\citep{kreiss2022concadia} and 
zero-shot (Socratic Model~\cite{zeng2022socratic}) baselines.

\textbf{Result}.
In Table~\ref{tab:concadia_table},
\modelname outperforms the Socratic Model~\cite{zeng2022socratic} implementation based on a stronger language model (GPT-3 175B~\cite{brown2020gpt3}).
Interestingly, the base VLM (BLIP-2) and \modelname (Ours) show a completely different text style. \modelname captions are better 
aligned with \textit{caption}-style, showing that our method reflects the Wikipedia article better than the baselines.
On the other hand, the VLM generates \textit{description}-style texts better.
Still, \modelname captions are similar to the visually intensive caption (\textit{description}) compared to all other baselines except for the VLM.

\subsection{Paragraph Captioning}
\label{subsec:exp_para}
Image Paragraph Captioning~\cite{krause2017imageparagraph}
has paragraph-long captions that describe the image in finer detail
than sentence-level captions.

\textbf{Approach and baselines}.
We saw that neither the VLM nor the text-only model could follow the style of the ground-truth annotation in early experiments.
Hence, we provide the model with three in-context examples (3-shot).
Note that the setting is still much more challenging compared to that of the fully supervised baselines
(Krause at el.~\cite{krause2017imageparagraph}, Liang et al.~\cite{liang2017recurrent}, SCST with repetition penalty~\cite{melas2018training}, HSGED~\cite{yang2020hierarchical}, and PaG-MEG-SCST~\cite{nguyen2022effective}).

\textbf{Results}.
As visible in Table~\ref{tab:paracap_table},
\modelname greatly improves the base VLM (BLIP-2)
to generate paragraph captions comparable to the supervised baselines.
We provide an interpretation of this improvement in qualitative samples in \cref{sec:ax_samples}:
\modelname shows less text degeneracy than the base VLM, while keeping visual hallucination at a minimum unlike \naive Ensemble.

\begin{table}
    \small
    \centering
    \setlength{\tabcolsep}{0.4em}
    \begin{tabular}{l|ccccc}
        Models & rep-2$\downarrow$ & div.$\uparrow$ & coh.$\uparrow$ & Mauve$\uparrow$ & CLIP.$\uparrow$ \\
        \hline
        Cont. Search & 2.60 & 0.97 & 0.34 & 0.86 & 0.65 \\
        MAGIC & 2.49 & 0.97 & 0.38 & 0.85 & 0.68  \\
        \hline
        BLIP-2 & 24.26 & 0.39 & 0.32 & 0.47 & \textbf{0.87}  \\
        \naive Ensemble & \textbf{1.85} & \textbf{0.98} & 0.27 & 0.93 & 0.67  \\
        \hline
        Ours & 2.31 & 0.97 & \textbf{0.38} & \textbf{0.96} & 0.72  \\
    \end{tabular}
    \caption{Results in the ROCStories story generation dataset~\cite{mostafazadeh2016rocstories}.
    \textit{rep-2} denotes 2-gram repetition, \textit{div.} diversity, \textit{coh.} coherence, and \textit{CLIP.} CLIPScore.
    Higher is better except for \textit{rep-2}.
    }
    \label{tab:story_table}
\end{table}

\subsection{Story Generation}
\label{subsec:exp_story}

Story generation is an open-ended generation task.
To excel at it, \modelname should generate open-ended text
without falling into text degeneracy, all the while staying close to the image context.

\textbf{Approach and baselines}.
Unlike previous experiments, here
we use a supervised text-only model~\cite{su2022a}
finetuned on text-only ROCStories~\cite{mostafazadeh2016rocstories} dataset.
Hence, we can safely assume that this specialist text-only model knows the language "better" than the VLM in story generation.
We include both visually-conditioned (MAGIC~\cite{su2022magic})
and text-only (Contrastive search~\cite{su2022contrastive})
baselines.
Refer to \cref{sec:ax_implementation} for more details on the baseline results.

\textbf{Results}.
Table~\ref{tab:story_table} 
presents the results of open-ended story generation.
\modelname outperforms both Contrastive Search and MAGIC in all metrics.
While \naive Ensemble builds more diverse text (rep-2 and div.), 
its severely low coherence score suggests that its stories are less consistent, as represented in qualitative samples of \cref{sec:ax_samples}.
Finally, while the base VLM (BLIP-2) shows high image-text correspondence as reflected in high CLIPScore,
it cannot generate an articulate story as its low performance on other scores shows.

\section{Qualitative Results}
\label{sec:qualitative}

\begin{figure}[t]
    \centering
    \includegraphics[width=0.48\textwidth]{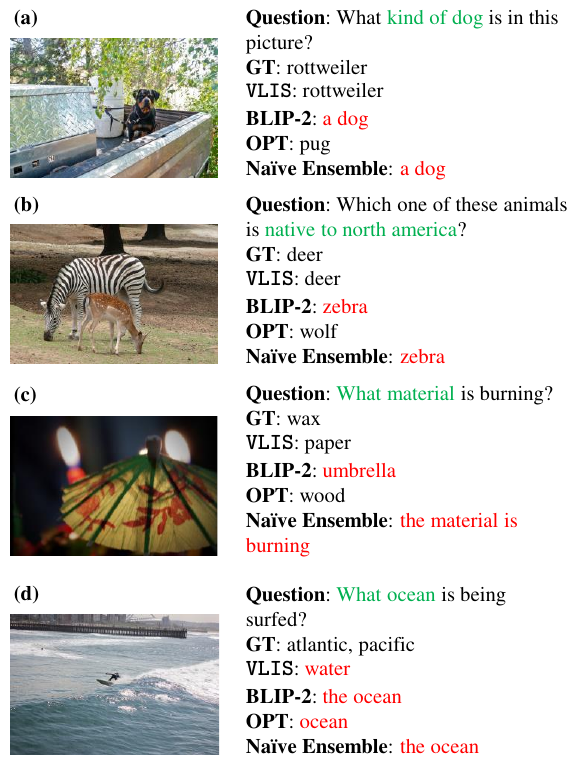}
    \caption{Generation results in the OK-VQA dataset~\cite{okvqa2019cvpr}. We color the intention of the question 
    \textbf{\textcolor{sgreen}{green}}
 and answers that defy such intention with \textbf{\textcolor{sred}{red}}.
    (c) and (d) are failure cases.}
    \label{fig:main_okvqa}
\end{figure}

\textbf{Commonsense Understanding}.
Figure~\ref{fig:main_okvqa} illustrates zero-shot results in the OK-VQA dataset~\cite{okvqa2019cvpr}.
In (a) and (b), the baselines including the base VLM and \naive Ensemble fail to understand the intention of the question (\textit{kind of dog} and \textit{native to north america}).
While the text-only model understands the question better and suggests plausible answer candidates (\textit{pug} and \textit{wolf}), it has no access to the visual inputs and ultimately outputs an incorrect answer.
On the other hand, \modelname sensibly combines commonsense reasoning and visual context.

\begin{figure*}[ht]
    \centering
    \includegraphics[trim=0cm 0.2cm 0cm 0cm,clip,width=0.99\textwidth]{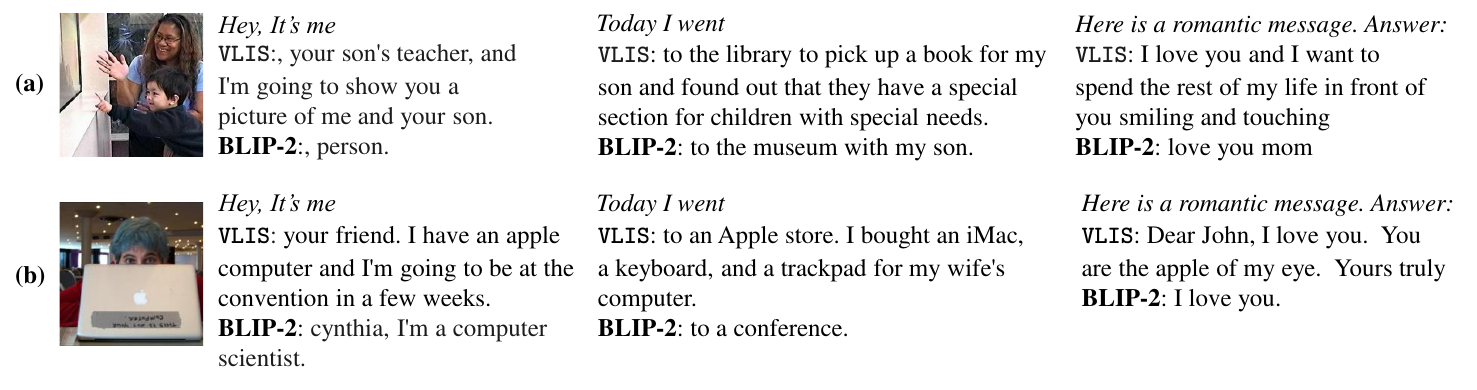}
    \caption{Open-ended generation results with BLIP-2~\cite{li2023blip} as the base VLM. We use three text prompts (\textit{Hey, It's me}, \textit{Today I went}, and \textit{Here is a romantic message. Answer:}) to test whether \modelname can actively adjust its response according to the text prompt while maintaining visual alignment.}
    \label{fig:main_openended}
\end{figure*}

Results for images (c) and (d) depict the failure cases. In (c), \modelname follows the reasoning process of the text-only language model to deduce that the answer should be a type of material. However, as the VLM focuses on the frontal object (\textit{umbrella}), \modelname wrongly concludes the answer is the material of that object (\textit{paper}, which is coincidentally a flammable material as well). In (d), the text-only model produces an incoherent output (\textit{ocean}). \modelname inherits this misinterpretation and likewise generates an incorrect answer (\textit{water}).
In conclusion, \modelname induces coordination of the VLM's visual specificity and the text-only model's commonsense understanding but carries on the modeling insufficiency of the individual modalities.


\textbf{Open-Ended Generation}.
Lastly, we demonstrate the open-ended generation capability of \modelname in Figure~\ref{fig:main_openended}.
Here, \modelname should condition its output on the diverse text prompt and the image.
Unlike the base VLM, it clings tighter to the prompt and produces realistic self-introduction (\textit{hey, it's me}), personal journal (\textit{today I went}), and romantic messages (\textit{here is a romantic message. answer:}).
Also, \modelname plays pun on the word \textit{apple}  
(see \textit{apple laptop} in the image and \textit{apple of my eye}).
Refer to \cref{sec:ax_samples} for more baseline samples.

\section{Related Work}
\label{sec:related_works}

\textbf{Combining VLMs with text-only LMs}.
Early large-scale VLMs (LXMERT~\cite{tan2019lxmert}, VisualBERT~\cite{li2019visualbert} and ViLBERT~\cite{lu2019vilbert}) saw the benefits of text-only pretraining by initializing their text encoder with a masked language model BERT~\cite{kenton2019bert}.
Later, Frozen~\cite{tsimpoukelli2021frozen} started a trend of freezing the language model and learning only the vision-language relationship.
More recent models such as Flamingo~\cite{alayrac2022flamingo} and BLIP-2~\cite{li2023blip} also freeze the image encoder.
ESPER~\cite{yu2022multimodal} uses reinforcement learning to combine image encoders with language models.

Better aligned with our approach are decoding-oriented methods for image-conditioned text generation. 
ZeroCap~\cite{tewel2022zerocap} uses the gradient signal from a pretrained image-text  alignment scorer (CLIP~\cite{radford2021clip}) to update the language model's memory.
Magic~\cite{su2022magic} also utilizes CLIP.
Unlike ZeroCap and Magic, \modelname utilizes autoregressive VLMs~\cite{li2023blip}, rather than CLIP.

\textbf{Language Model Decoding}.
Language model decoding is the process of generating text from a pretrained language model.
Traditional decoding methods use greedy decoding and beam search to find the most likely sequence of words.
The truncated sampling algorithms such as
Top K sampling~\cite{fan2018hierarchical,holtzman2018learning,radford2019language}, Nucleus sampling~\cite{holtzman2020curious}, and Typical P sampling~\cite{Meister2022LocallyTS}
have been proposed to avoid text degeneracy.
Recent deterministic algorithms, such as 
Contrastive decoding~\cite{li2022contrastive} and contrastive search~\cite{su2022a,su2022contrastive}, provide a better trade-off between text fluency and model likelihood.
Neurologic~\cite{lu2021neurologic} and Neurologic A*esque decoding~\cite{lu2022neurologic} control the language models to include given words in their outputs.
Concurrently with our work, ZEROGEN~\cite{tu2023zerogen} extends the task of controllable text generation to the multimodal domain.
As shown in the experiments, \modelname can be used jointly with any decoding method, including beam search and contrastive search.

\section{Conclusion}
\label{sec:conclusion}

We propose \modelname, a novel framework
to alleviate the language modeling burden of visual-language models (VLMs).
\modelname combines the linguistic understanding capability of the text-only language models with the visual conditioning strength of the VLMs by importance sampling.
To isolate the VLMs' visual conditioning power,
\modelname uses pointwise mutual information to suppress their text-only marginal distribution.
Our framework enhances the base VLM in commonsense reasoning (WHOOPS~\cite{bitton2023whoops}, OK-VQA~\cite{okvqa2019cvpr}, and ScienceQA~\cite{lu2022scienceqa})
and complicated text generation
(Concadia~\cite{kreiss2022concadia}, Image Paragraph Captioning~\cite{krause2017imageparagraph},
 and ROCStories~\cite{mostafazadeh2016rocstories}) problems.
In the future,
\modelname can be extended 
to incorporate other modalities for which the paired multimodal data is even scarcer.
We hope that \modelname sparks an interest in better utilization of off-the-shelf multimodal pretrained models.


\section{Ethical Considerations \& Limitations}
\label{sec:limitations}

\textbf{Potential ethical concerns}.
As an inference time method, \modelname inherits some known problems of both the VLMs and the unimodal text-only language models as well:

\begin{itemize}
    \item Hallucination: VLMs are known to hallucinate information absent in the training data~\cite{rohrbach-etal-2018-hallucination}. While \modelname may strengthen visual conditioning and thereby contribute to reducing the rate of visual hallucination, completely eradicating it is beyond the scope of this research.
    \item Social bias: It is widely known that VLMs reflect or even amplify~\cite{hendricks2018women,hirota2022quantifying} social bias (\eg gender or race) in the training data. We have yet to determine how \modelname affects social bias in the base models. Thus, outputs generated using \modelname may contain social bias.
\end{itemize}

It is a meaningful direction to combine \modelname with reinforcement learning~\cite{Ramamurthy2022IsRL, Yu_2023_CVPR} or reward-based decoding algorithm~\cite{su2022magic} to alleviate the problems above, but we leave that to future research.

\textbf{Limitation of \modelname and future work}.
Firstly, we acknowledge that this paper only explores a small fraction 
of the possible combinations of text-only models and VLMs.
A large-scale wide search in this regard would reveal 1) the better-performing pairs of text-only LM and VLM and
2) the required characteristics of a good model pair.

Secondly, \modelname could be extended to more modalities than the image-to-text generation problem covered here.
Other modalities, such as audio and document may also benefit from applying \modelname to their modality-specific foundational model.

Finally, \modelname can be short-sighted. The method combines the outputs of the VLMs and the text-only models at the very last stage of token likelihood. As a result, \modelname score might be misleading when both models assign high probabilities to the same token for different reasons (\eg homophones). It may help to estimate scores for the future generated text by rolling out a few generative steps and aggregating the output~\cite{lu2022neurologic}, which we leave to future works.

\section{Acknowledgement}
\label{sec:acknowledgement}

This work was supported by Institute of Information \& communications Technology Planning \& Evaluation (IITP) grant funded by the Korea government (MSIT) (No.2020-0-01361), Institute for Project-Y, and NCSOFT Vision/NLP Center.

\bibliography{main}
\bibliographystyle{acl_natbib}

\clearpage

\appendix

\section{VLM Failure Cases}
\label{sec:ax_failure}

\subsection{Landmark Recognition Experiment}
\label{subsec:ax_landmark}

To better understand the named entity recognition problem in VLMs' image descriptions,
we check whether their descriptions for pictures of popular landmarks contain the proper names.
We first collect the names of the 100 most popular landmarks~\footnote{The list for landmarks is from \url{www.listchallenges.com/100-most-famous-landmarks-around-the-world}.}.
Then, we filter the list by removing names of landmarks without proper nouns (\eg \textit{Middle of the Earth}), keeping 80 landmarks in total.
Finally, we download the corresponding pictures from Wikipedia.
Given the prompt \textit{What is this?}, we task the VLM to generate a response as long as 100 tokens and check whether the output contains the name of the given landmark.
Note that some landmarks have alternative names. Hence, we collect alternative names from Wikipedia and count the model-generated answer as correct when it contains any of the possible names.
Finally, we check whether the model tried to answer or not by inspecting whether the model-generated text contains the name of any landmark in our list.
We calculate the precision score by dividing the number of correct predictions by the number of tries.

Our landmark dataset~\footnote{We will release the dataset to the public.} is tiny compared to the similar dataset~\cite{weyand2020googlelandmark} for a purpose: we want to check whether the VLM avoids telling the named entities, not whether the VLM saw them in the training process.
Hence, we narrow the scope of evaluation to the most popular landmarks,
in which we can assume that most of the entity names are found in the VLM training dataset.

\begin{figure}[ht]
    \centering
    \includegraphics[width=0.48\textwidth]{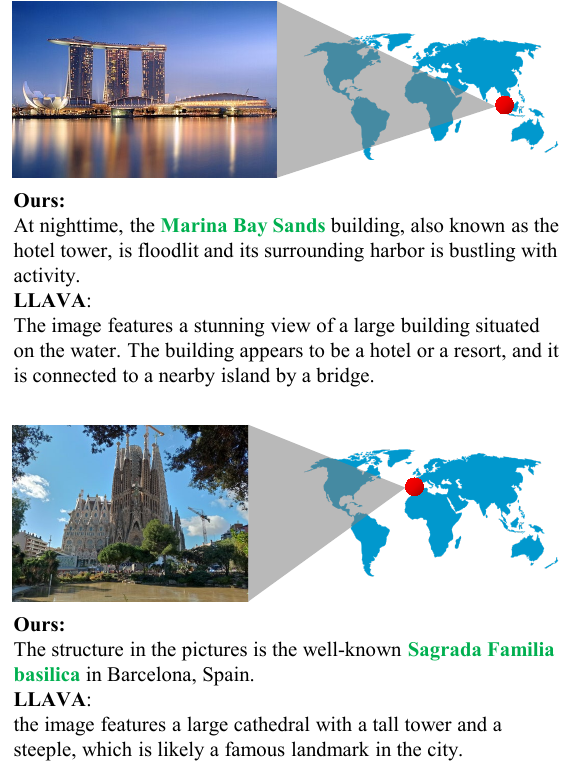}
    \caption{Comparison of LLAVA and \modelname in the landmark recognition experiment.}
    \label{fig:ax_landmark}
\end{figure}

Table~\ref{tab:landmark_table} and Figure~\ref{fig:ax_landmark} compare base LLAVA~\cite{liu2023llava} and \modelname in our landmark recognition dataset.
The result shows that the VLM (LLAVA) knows at least about half the landmarks' names,
but does not tell them without applying \modelname.
Also, \modelname shows good precision, showing that
it does not get more correct answers by guessing more.
We further demonstrate that a proper answer to our prompt \textit{What is this?} should contain the name of the landmarks:
when we present GPT3 with the ground-truth alt captions and the prompt,
GPT3 always includes the landmark names in its output.

\begin{table}
    \centering
    \begin{tabular}{l|c|cc}
        Models & GT Caption & Acc & Prec \\
        \hline
        GPT3 & \checkmark & 1.00 & 1.00 \\
        \hline
        LLAVA & & 0.16 & 0.48 \\
        Ours & & \textbf{0.41} & 0.70
    \end{tabular}
    \caption{Results on our landmark recognition experiment. \textit{Acc} denotes accuracy and \textit{Prec} denotes precision.}
    \label{tab:landmark_table}
\end{table}

\subsection{More Qualitative Results}
\label{subsec:ax_failure_samples}

\begin{figure*}
    \centering
    \includegraphics[width=0.99\textwidth]{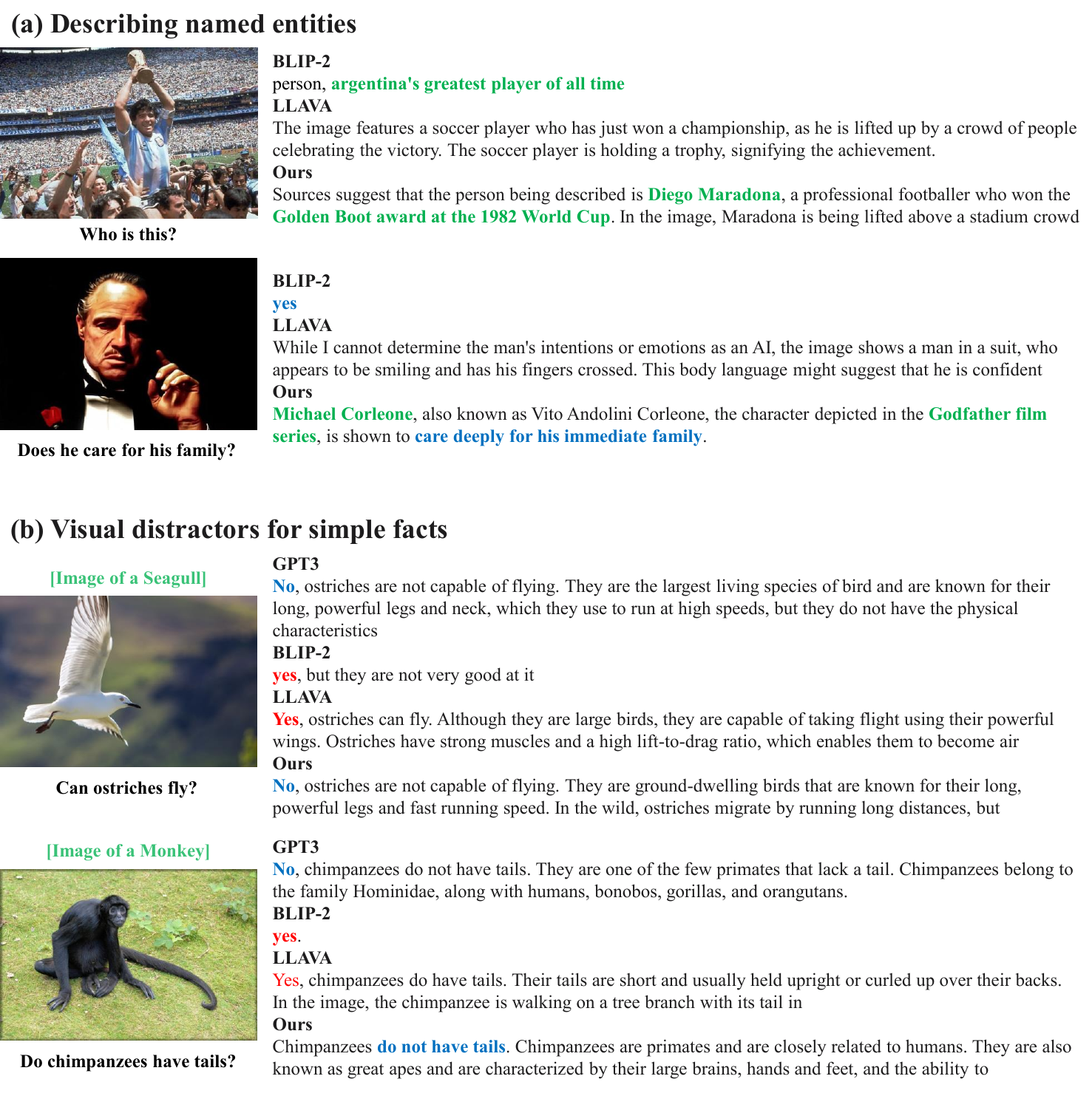}
    \caption{Raw text output for samples described in Figure~\ref{fig:main_intuition}.}
    \label{fig:ax_intuition}
\end{figure*}

\begin{figure*}
    \centering
    \includegraphics[width=0.99\textwidth]{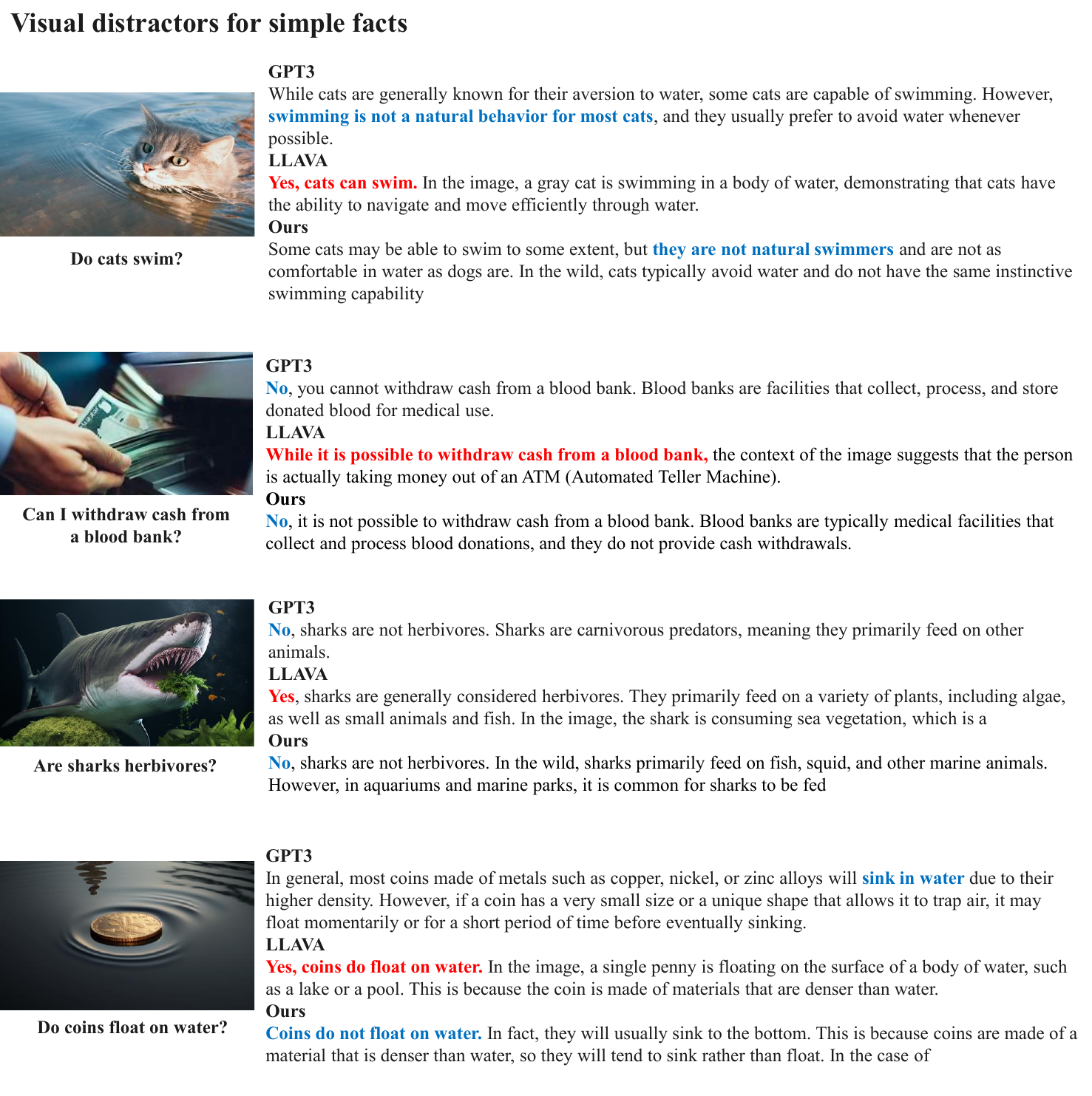}
    \caption{More examples in which the visual context deceives the base VLM (LLAVA~\cite{liu2023llava}) into denying simple facts.}
    \label{fig:ax_distractor}
\end{figure*}

Figure~\ref{fig:ax_intuition} shows full raw text outputs for the VLM failure cases shown in Figure~\ref{fig:main_intuition}.
Figure~\ref{fig:ax_distractor} depicts more samples for the failure case 2:
the base VLM (LLAVA) is distracted by misleading visuals while \modelname does not.

\section{Implementation Details}
\label{sec:ax_implementation}

\textbf{Computational Requirements}.
Using LLM.int8 approximation~\cite{dettmers2022llmint8}, a single NVIDIA TITAN RTX GPU (24GB Memory) fits both the BLIP-2 2.7B and OPT 1.3B models. Flan-T5 XL and XXL models need more memory and \modelname using the larger backbones requires NVIDIA A6000 GPU (48GB) for inference.
Both LLAVA 13B and Vicuna 7B fit into an A6000 GPU at the same time.
Generating 50 tokens takes $\sim 20$ seconds in all settings.

\textbf{Hyperparameters}.
We fix the fluency threshold $\alpha = 0.001$ in all experiments and use beam search with beam size $5$. For QA problems, we apply length penalty $<0$ on the beam score
to induce succinct answers following the literature~\cite{li2023blip}.
The opposite behavior is required for longer text generation, so we set the value larger than $0$ for open-ended generation problems.
The language temperature $\tau$ is manually selected by examining the text quality of three samples per task.

\textbf{Task-Specific Hyperparameters}.
For VQAv2~\cite{goyal2017vqav2}, OK-VQA~\cite{okvqa2019cvpr}, and ScienceQA~\cite{lu2022scienceqa} datasets, we set the language temperature $\tau=1.25$ and length penalty $-1.0$ to
induce succinct answers generated with stronger visual conditioning.
In Concadia~\cite{kreiss2022concadia}, $\tau=0.67$ and length penalty $-2.0$ is used for succinct caption-style text with better text conditioning.
For Image Paragraph Captioning~\cite{krause2017imageparagraph} experiments we use $\tau=0.67$ and length penalty $1$ to induce longer captions. Also, we apply contrastive search~\cite{su2022contrastive} with a penalty of $0.6$ to avoid text degeneracy.

\textbf{Flan-T5 Hyperparameters}.
For the backbone comparison study in \cref{sec:ax_backbones}, we set the VLM backbone to BLIP-2 Flan-T5~\cite{li2023blip} and text-only model to Flan-T5~\cite{chung2022flan}. For Flan-T5 variants, we compensate the overconfidence of the model with a large temperature of $1.5$ to normalize the logit outputs. For the same reason, we also relax the fluency threshold $\alpha = 0.0001$. 
Finally, the language temperature $\tau$ is set to $0.9$.

\textbf{Baseline Hyperparameters}.
We share the same hyperparameters as in \modelname for all our implemented baselines; LLAVA, BLIP-2, OPT-IML, and \naive Ensemble. We do not modify the beam size $5$ and fluency threshold $\alpha = 0.001$,
and change the length penalty accordingly to the task following the \modelname hyperparameters.

\textbf{Few-Shot Settings}.
For Image Paragraph Captioning~\cite{krause2017imageparagraph}, we use three ground-truth examples to prime the models for the paragraph-long generation task.
However, one cannot provide multiple images as inputs to the backbone VLM model (BLIP-2~\cite{li2023blip}).
Hence, we simply insert the few-shot samples in the text domain and provide only the single target image as the visual context.

\textbf{uint8 Inference}.
LLM.int8~\cite{dettmers2022llmint8} is an approximated inference technique for large language models. It applies vector-wise quantization and mixed-precision decomposition to reduce memory consumption without performance degradation.
We employ the technique to jointly run both text-only LM and VLM on a single GPU.

\textbf{Randomness}.
As \modelname is a deterministic inference time algorithm, no randomness is involved in any of the experiments. A stochastic sampling version of \modelname may require variance analysis, but we leave that to future research.

\textbf{Evaluating Story Generation}.
While the official repository of MAGIC~\cite{su2022magic} shares the inference results, it does not contain the evaluation scripts. Thus, we consult the repository Contrastive Decoding~\cite{li2022contrastive} for the evaluation script for an open-ended generation problem.
Due to the difference in the evaluation code, our baseline scores are different from the results reported in MAGIC~\cite{su2022magic}.
However, we still use the public inference results for the baselines and evaluate each model with a publicly available code, making our evaluation pipeline unbiased, transparent, and reproducible.

\begin{table}[t]
    \centering
    \begin{tabular}{p{1.8cm}|cc|c}
        Models & Random & \# Images & OKVQA \\
        \hline
        VLM-only & & & 31.7 \\
        \hline
        Ours & \textbf{False} & \textbf{2} & \textbf{34.2} \\
        \hline
        Ours & True & 1 & 29.0 \\
        Ours & True & 2 & 32.2 \\
        Ours & True & 10 & 35.3
    \end{tabular}
    \caption{Results in the OK-VQA validation set. Our default option (prefined set with two images) is marked bold.}
    \label{tab:ax_marginal}
\end{table}

\section{Marginal Approximation Experiment}
\label{sec:ax_marginal}

In the main paper, we propose using one or two images with minimal visual information (black-filled and white-filled) as a functional candidate with minimum computational overhead. 
To investigate the alternative approaches, we conducted an additional experiment in the OK-VQA dataset.
The variables considered here are 1. Random vs. predefined (black-filled and white-filled) set of images and 2. The number of images used to approximate the expectation. We keep everything else the same as in Table~\ref{tab:okvqa_table} and only adjust the marginal approximation scheme.

Our results are summarized in Table~\ref{tab:ax_marginal}.
First, a random set of images is inferior to our predefined set of images for approximating the marginal.
Second, 10 random image set offers a better approximation than the predefined set of two images. Still, the 10 random images option requires 11 passes of VLM per token generation, making it largely inefficient for practical usage.

\begin{table}[t]
    \centering
    \begin{tabular}{p{1.8cm}|c|c}
        Models & $\alpha$ & OKVQA \\
        \hline
        VLM-only & & 31.7 \\
        \hline
        Ours & 1e-1 & 13.8 \\
        Ours & 1e-2 & 30.1 \\
        \hline
        Ours & \textbf{1e-3} & \textbf{34.2} \\
        \hline
        Ours & 1e-4 & 34.4 \\
        Ours & 1e-5 & 33.1 \\
        Ours & 0 & 32.3
    \end{tabular}
    \caption{Results in the OK-VQA validation set. Our default fluency threshold value ($\alpha=1e-3$) is marked bold.}
    \label{tab:ax_fluency}
\end{table}

\section{Fluency Threshold Experiment}
\label{sec:ax_fluency_threshold}

Here, we examine the effect of fluency threshold value $\alpha$
on the generation quality of \modelname.
This experiment extends the OK-VQA commonsense reasoning experiment in Table~\ref{tab:okvqa_table} and keeps all other variables the same except for $\alpha$.

Table~\ref{tab:ax_fluency} shows that \modelname consistently outperforms the VLM-only baseline
for all values of $\alpha$ in the range of $[1e-3, 1e-5]$.
Too large values ($[1e-1, 1e-2]$) still harm the performance as they typically leave only one or two token candidates for the \modelname Score to choose from.

\section{Backbone Scale Experiment}
\label{sec:ax_backbones}

We conduct a comparison study to 
test whether the improvement offered by \modelname
is generalizable to a wider set of architectures and model sizes.
Here, we mainly evaluate \modelname with Flan-T5 variants as both the text-only LM and VLM backbones.
T5~\cite{raffel2020t5} is an encoder-decoder transformer unlike the decoder-only autoregressive language models (\eg OPT~\cite{zhang2022opt} and GPT-3~\cite{brown2020gpt3}).
Flan-T5~\cite{chung2022flan} further trains T5 for better responsiveness in instruction prompts.
Table~\ref{tab:ax_scaling_table} summarizes the backbone comparison results on the OK-VQA dataset~\cite{okvqa2019cvpr}.
In all combinations of model sizes except for FlanT5\textsubscript{Base}, \modelname improves the commonsense reasoning capability of the VLM backbone.
Also, \naive Ensemble performs unreliably depending on the choice of the text-only LM and performs worse than the VLM itself in most of the settings.
The FlanT5\textsubscript{Base} LM makes \modelname work worse than the VLM. Since \modelname is built on the assumption that the text-only LM knows the human language distribution better than the VLM, this deterioration of performance further supports our explanation of why \modelname works.

\begin{table}
    \centering
    \small
    \setlength{\tabcolsep}{0.5em}
    \begin{tabular}{ll|cccc}
        VLM & LLM & \multirow{2}{*}{Ours} & \multicolumn{2}{c}{Vanilla} & \naive \\
        Backbone & Backbone &  & LLM & VLM & Ensemble \\
        \hline
        OPT\textsubscript{2.7B} & OPT\textsubscript{1.3B} & 34.2 & 19.1 & 31.7 & 26.6 \\
        \hline
        F-T5\textsubscript{XL} & F-T5\textsubscript{Base} &  29.8 & 12.5 & 40.7 & 34.4  \\
        F-T5\textsubscript{XL} & F-T5\textsubscript{XL} &  43.4 & 19.3 & 40.7 & 39.0 \\
        F-T5\textsubscript{XL} & F-T5\textsubscript{XXL} & 43.9 & 21.3 & 40.7 & 42.0 \\
        F-T5\textsubscript{XXL} & F-T5\textsubscript{XXL} & 47.5 & 21.3 & 45.9 & 44.4
    \end{tabular}
    \caption{Backbone comparison experiments on the validation set of the OK-VQA dataset~\cite{okvqa2019cvpr}. F-T5 denotes T5 trained on FLAN dataset~\cite{weifinetuned}.}
    \label{tab:ax_scaling_table}
\end{table}

\section{Prompt Templates}
\label{sec:ax_templates}

In the prompt templates below,
TLM denotes the prompt presented to the text-only model
and VLM denotes that given to the VLM.

\begin{itemize}
    \item  \textbf{OK-VQA \& VQAv2}
\begin{itemize}
\item Variables: 
 \begin{spverbatim}[QUESTION] \end{spverbatim}
    \item TLM
\begin{spverbatim}
Question: [QUESTION] Answer:
\end{spverbatim}
    \item VLM
\begin{spverbatim}
Question: [QUESTION] Answer:
\end{spverbatim}
\end{itemize}
\item  \textbf{ScienceQA}
\begin{itemize}
\item Variables:
 \begin{spverbatim}[QUESTION], [CONTEXT], [CHOICES] \end{spverbatim}
\item TLM
\begin{spverbatim}
Answer the multi-choice question given the image. Question: [QUESTION] [CONTEXT] Choices: [CHOICES] Answer:
\end{spverbatim}
    \item VLM
\begin{spverbatim}
Answer the multi-choice question given the image. Question: [QUESTION] [CONTEXT] Choices: [CHOICES] Answer:
\end{spverbatim}
\end{itemize}
\item  \textbf{Concadia}
\begin{itemize}
    \item TLM
\begin{spverbatim}
Write a short caption that describes the image. Article: "[ARTICLE]" Caption:
\end{spverbatim}
    \item VLM
\begin{spverbatim}
The image describes
\end{spverbatim}
\end{itemize}

\item  \textbf{Image Paragraph Captioning}
\begin{itemize}
\item Variables: \begin{spverbatim}[ARTICLE] \end{spverbatim}
    \item TLM
\begin{spverbatim}
Write a multi-sentence long paragraph describing the image. Each sentence should describe a different aspect of the image and should be concise.\n
(Image 1) Image Description: [Description Sample 1]\n
(Image 2) Image Description: [Description Sample 2]\n
(Image 3) Image Description: [Description Sample 3]\n
(Image 4) Image Description:
\end{spverbatim}
    \item VLM
\begin{spverbatim}
Write a multi-sentence long paragraph describing the image. Each sentence should describe a different aspect of the image and should be concise.\n
(Image 1) Image Description: [Description Sample 1]\n
(Image 2) Image Description: [Description Sample 2]\n
(Image 3) Image Description: [Description Sample 3]\n
(Image 4) Image Description:
\end{spverbatim}
\end{itemize}

\item \textbf{ROCStories}
\begin{itemize}
    \item Variables:
 \begin{spverbatim}[TOPIC] \end{spverbatim}
\item TLM
\begin{spverbatim}
    [TOPIC] <|endoftext|>
\end{spverbatim}
\item VLM
\begin{spverbatim}
    The image describes
\end{spverbatim}
\end{itemize}

\item \textbf{WHOOPS}
\begin{itemize}
\item TLM
\begin{spverbatim}
Decide whether the image is strange or natural in terms of physics, commonsense, or etc.\n
Start with "The image shows"
\end{spverbatim}
\item VLM
\begin{spverbatim}
Decide whether the image is strange or natural in terms of physics, commonsense, or etc.\n
Start with "The image shows"
\end{spverbatim}
\end{itemize}

\end{itemize}

\section{More Qualitative Samples}
\label{sec:ax_samples}

\begin{figure*}
    \centering
    \includegraphics[width=0.99\textwidth]{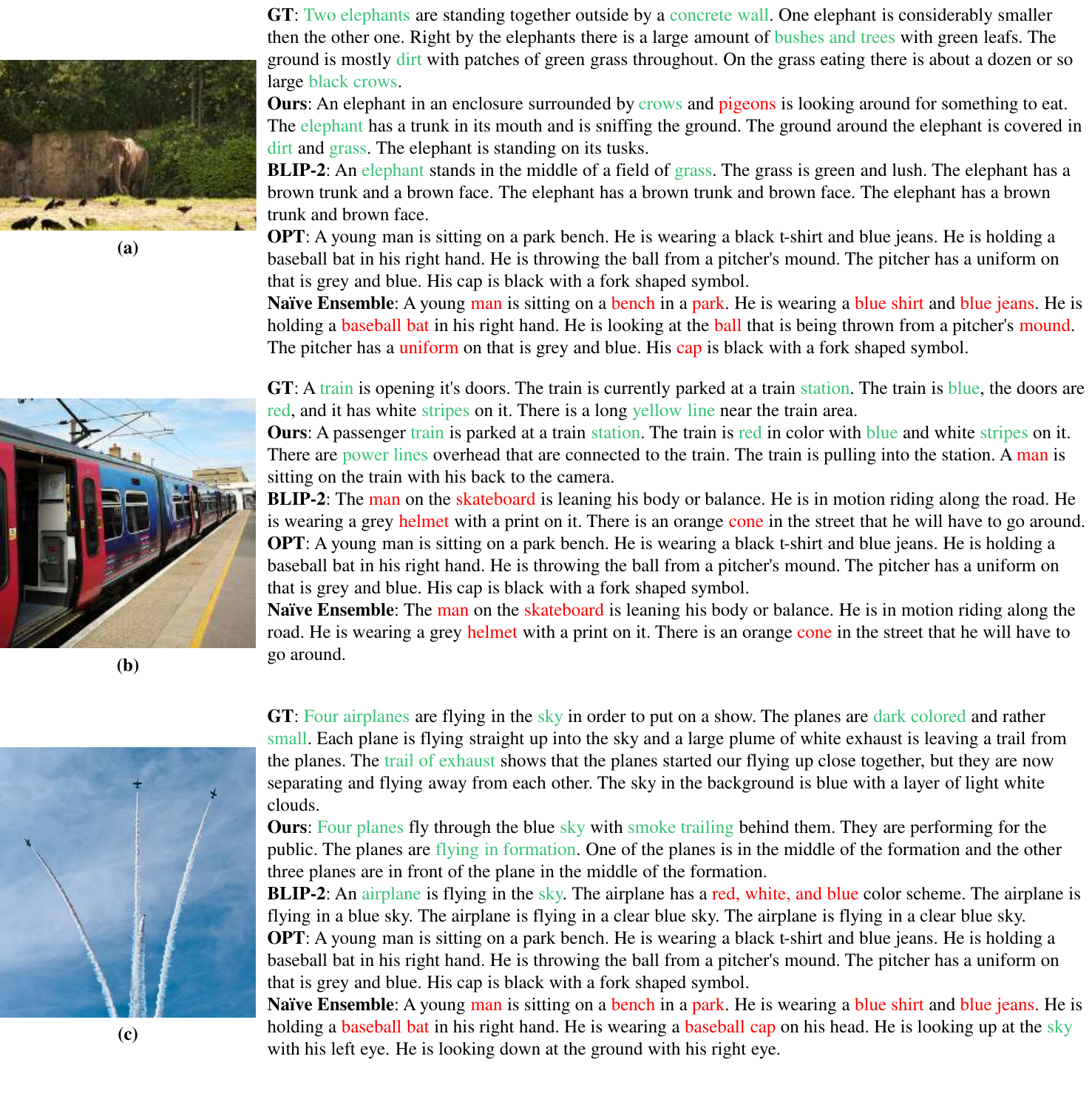}
    \caption{Generation results on the Image Paragraph Captioning dataset~\cite{krause2017imageparagraph}.
    The correct visual expressions are marked in 
    \textbf{\textcolor{sgreen}{green}}
    and wrong ones are in \textbf{\textcolor{sred}{red}}.
    }
    \label{fig:ax_para}
\end{figure*}

\begin{figure*}
    \centering
    \includegraphics[width=0.99\textwidth]{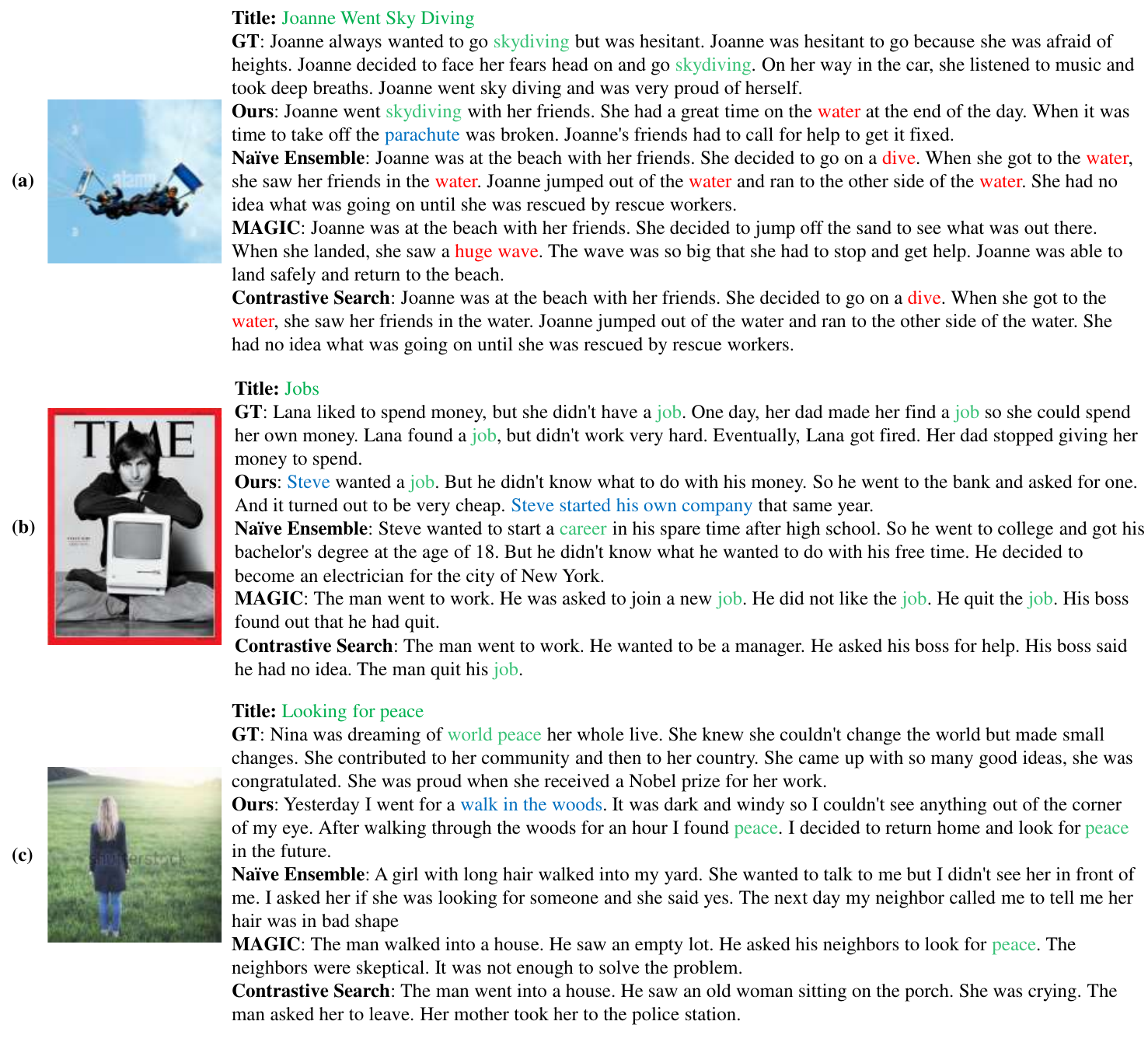}
    \caption{Generation results on the ROCStories dataset~\cite{mostafazadeh2016rocstories}.
    Expressions showing alignment to the title are colored in \textbf{\textcolor{sgreen}{green}},
    alignment to the image in \textbf{\textcolor{sblue}{blue}},
    and misinterpretations in \textbf{\textcolor{sred}{red}}.}
    \label{fig:ax_story}
\end{figure*}

\begin{figure*}
    \centering
    \includegraphics[width=0.99\textwidth]{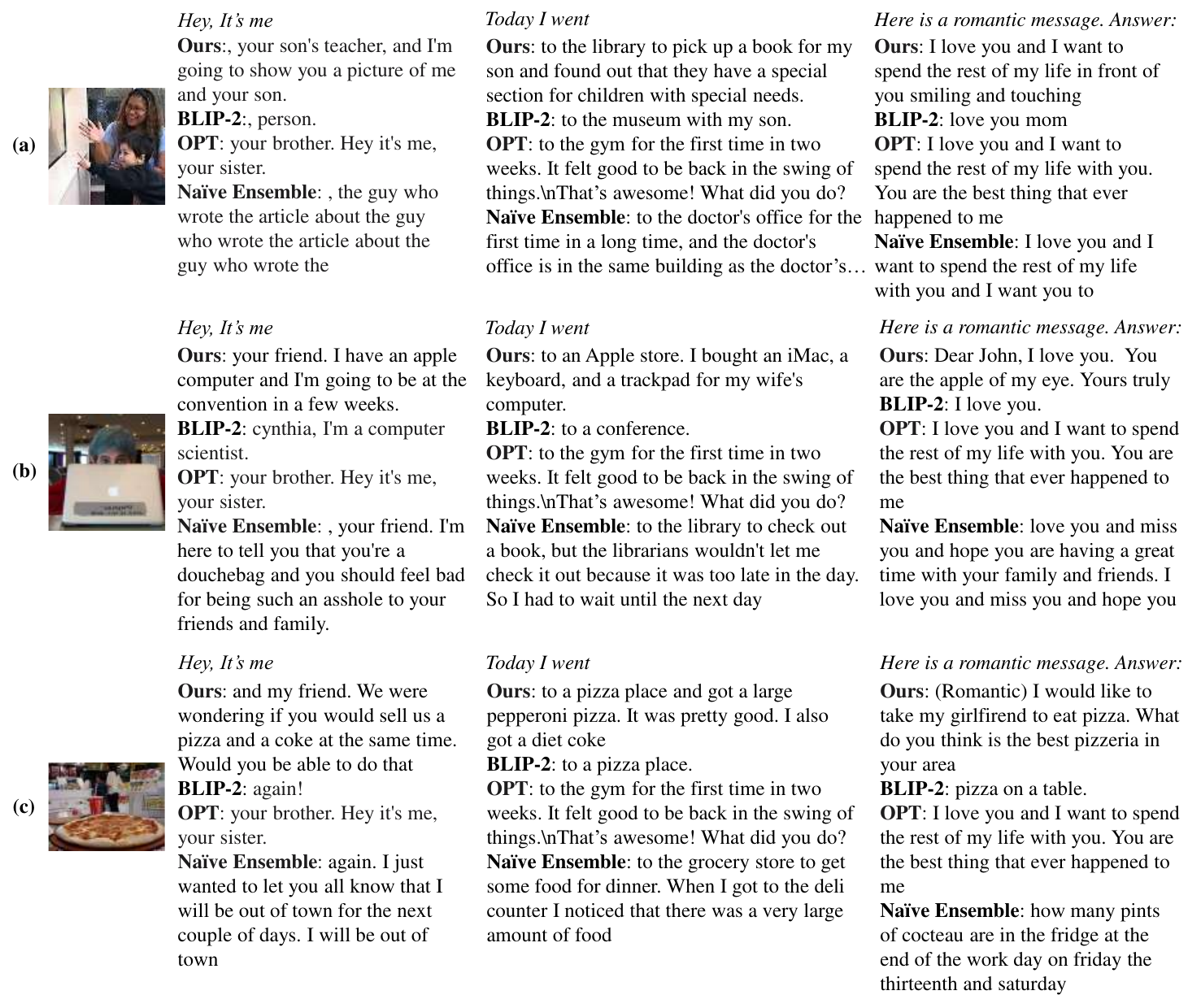}
    \caption{Open-ended generation results with various text prompt.
    Here we include more baselines than in Figure~\ref{fig:main_openended}.}
    \label{fig:ax_open_ended}
\end{figure*}

We include more qualitative samples
for image paragraph generation in Figure~\ref{fig:ax_para},
story generation in Figure~\ref{fig:ax_story},
and open-ended generation in Figure~\ref{fig:ax_open_ended}.


\end{document}